\documentclass[letterpaper, 10 pt, conference]{IEEEtran}  

\usepackage{graphics} 
\usepackage{epsfig} 
\usepackage{mathptmx} 
\usepackage{times} 
\usepackage{amsmath} 
\usepackage{amssymb}  
\usepackage{multirow}
\usepackage[ruled,vlined]{algorithm2e}
\usepackage{lipsum}
\usepackage{hyperref}
\usepackage{adjustbox}
\usepackage{draftwatermark}
\SetWatermarkText{Submitted to IEEE SMC 2019}
\SetWatermarkScale{0.15}
\usepackage{subfigure}
\usepackage{xcolor}
\hypersetup{
	colorlinks,
	linkcolor={red!50!black},
	citecolor={blue!50!black},
	urlcolor={blue!80!black}
}
\newcommand*{\affaddr}[1]{#1} 
\newcommand*{\affmark}[1][*]{\textsuperscript{\dag}}
\newcommand*{\email}[1]{\texttt{#1}}

\SetAlFnt{\small}
\title{\LARGE \bf
Sensing Volume Coverage of Robot Workspace using On-Robot Time-of-Flight Sensor Arrays for Safe Human Robot Interaction
}
\author{%
	Shitij Kumar\affmark[1], Ferat Sahin\affmark[1]\\
	\affaddr{\affmark[1]Department of Electrical and Microelectronic Engineering}\\
	\email{\{spk4422\textsuperscript{*},feseee\}@rit.edu}\\
	\affaddr{Rochester Institute of Technology, Rochester NY 14623, USA}%
}

\begin{document}

\maketitle
\thispagestyle{empty}
\pagestyle{empty}

\begin{abstract}

In this paper, an analysis of the sensing volume coverage of robot workspace as well as the shared human-robot collaborative workspace for various configurations of on-robot Time-of-Flight (ToF) sensor array rings is presented. A methodology for volumetry using octrees to quantify the detection/sensing volume of the sensors is proposed. The change in sensing volume coverage by increasing the number of sensors per ToF sensor array ring and also increasing the number of rings mounted on robot link is also studied. Considerations of maximum ideal volume around the robot workspace that a given ToF sensor array ring placement and orientation setup should cover for safe human robot interaction are presented. The sensing volume coverage measurements in this maximum ideal volume are tabulated and observations on various ToF configurations and their coverage for close and far zones of the robot are determined. 

\end{abstract}




\section{Introduction}

For a safe human robot interaction to occur, the robot must possess the information associated with its environment via exteroceptive sensors such as lidar(s), cameras and radars \cite{siciliano2016springer}. The placement of these sensors in the environment, determine the sensing volume coverage of the robot workspace. Using sensors that are mounted on the robot can provide information from the its perspective whilst removing the constraints of planning the placement of the sensors in the environment. They can also provide direct observations without the need of applying transformations to elicit relevant distance information associated with the human. This was implemented in our previous work \cite{kumarDynamicAwarenessIndustrial2018} and \cite{CASE2019_paper}. 

One of the ways of maintaining the safety of a human operator during human-robot interaction is speed and separation monitoring methodology (SSM) \cite{ISOTS15066}. To achieve SSM, the minimum separation distance and relative velocities between the robot and the human must be determined. The previous work \cite{CASE2019_paper} shows the implementation of a SSM safety configuration using three sensor arrays/rings consisting of eight Time-of-Flight laser-ranging sensors (also known as single-unit solid state lidar(s)) fastened around the robot links as shown in Figure 1.
\begin{figure}[h!]
    \centering
    \includegraphics[width=0.4\textwidth,keepaspectratio]{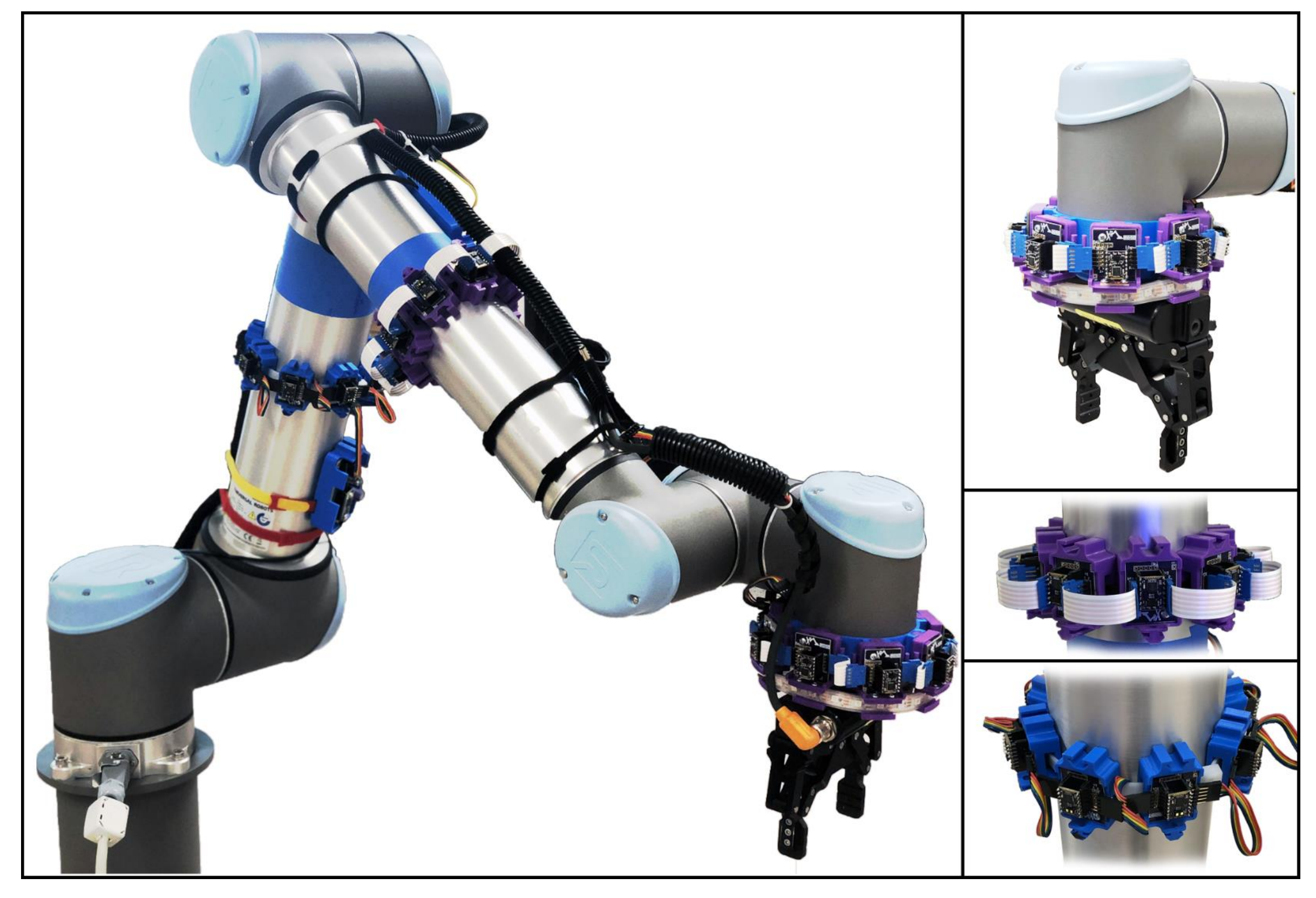}
    \caption{The UR10 robot with time-of-flight sensor arrays mounted on the robot link centers. Each array has eight single unit lidar(s) This sensing system is modelled in simulation to determine its sensing coverage. \cite{CASE2019_paper}}
    \vspace{-1.55em}
    \label{fig:TOFSetupOnRobot}
\end{figure}


Each array is considered as an augmentation to the robot body such that each observation incoming from an array is interpreted as an extension of the kinematic chain of the robot. This enables the sensing strategy to leverage the robot motion and provide exclusive coverage from the areas in the workspace where the robot is, and headed to. The setup also allows flexibility in terms of on-robot placement; the arrays can positioned anywhere on the robot links to achieve an optimal sensing coverage. Unlike the 2D scanning lidars, that provide planer information of separation and relative speed, this approach provides a 3D information with respect to the robot joint positions. 

Determining the minimum distance between two bodies is a non-trivial problem and central to any speed and separation monitoring methodology. In order to ensure correct and accurate measurements, the ToF sensor arrays should have the ability to sense a volume of the robot workspace. This study analyses the volume coverage of the ToF sensor arrays and presents a methodology to calculate sensing volume using octree based volumetry \cite{Octreeszeliski1993rapid}. 


Using off-robot sensors that are positioned around the robot or on the human operator to maximize coverage of sensors in the workspace has been the focus of many recent works. Recently, a 2D Lidar was used in conjunction with an IMU based human motion tracking setup \cite{safeeaMinimumDistanceCalculation2019}. In \cite{flaccoDepthSpaceApproach2015}, the authors used RGB-D cameras and proposed a novel approach to compute minimum distances in depth space instead of the Cartesian space and also introduced the idea of robot body approximation using few key-points. We rely on the contribution of the aforementioned and \cite{marvelPerformanceMetricsSpeed2013}, where the authors provided metrics for speed and separation monitoring, with two 2D lidar(s) that were used to track the human position with respect to a suspended manipulator. 

All the approaches mentioned so far have exclusively used proximity or inertial based sensing modalities and approaches, extrinsic to the robot. However, in \cite{schleglVirtualWhiskersHighly2013}, the authors introduced a new type of intrinsic perspective capacitive sensor that encouraged close operation between the human and the robot. In \cite{cerianiOptimalPlacementSpots2013} and \cite{lacevicKinetostaticDangerField2010}, the authors assessed the placement and orientation of IR distance sensors on a robot manipulator and implemented a kineostatic safety assessment algorithm, respectively. A reactive collision avoidance strategy was also implemented in \cite{RoccoABBSensors} using on-robot proximity sensors. In \cite{lacevicKinetostaticDangerField2010} and \cite{cerianiOptimalPlacementSpots2013}, authors used distance sensors for potential fields and tested the sensors placement on the robot body to examine the sensing volume coverage of the work space. In the work \cite{OperationAreaPointCloud}, the authors segment the volume of operating workspace using point-clouds with application in safe human robot interaction. Unlike the work in \cite{cerianiOptimalPlacementSpots2013}, where infrared distance (IR) sensors were placed individually, in this work Time-of-Flight sensor arrays mounted on robot links as `rings' were implemented \cite{kumarDynamicAwarenessIndustrial2018}.

This work analyzes the sensing volume coverage of robot workspace as well as the shared human-robot collaborative workspace for various configurations of ToF rings. It presents a methodology for volumetry using octrees to quantify the detection/sensing volume of the sensors, and how its coverage can be used to determine the choice of placement of ToF rings based on the task specific human robot interaction.
The remainder of the paper is organized as follows: Section \ref{sec:approach} describes the methodology for calculating sensing volume coverage for Time-of-Flight (ToF) range sensor configurations. The experiment setup is described in Section \ref{sec:experiment_validation}. The results of the experiments are shown and discussed in Section \ref{sec:Results}. Conclusions are drawn and future work discussed in Section \ref{sec:Conclusion}.

\section{Methodology}
\label{sec:approach}

\subsection{Time-of-Flight Sensor Array Setup}
\label{sec:TOFSetup}
\begin{figure}
		\centering
	\begin{subfigure}[]{
			\centering
			\includegraphics[width=0.45\linewidth,keepaspectratio]{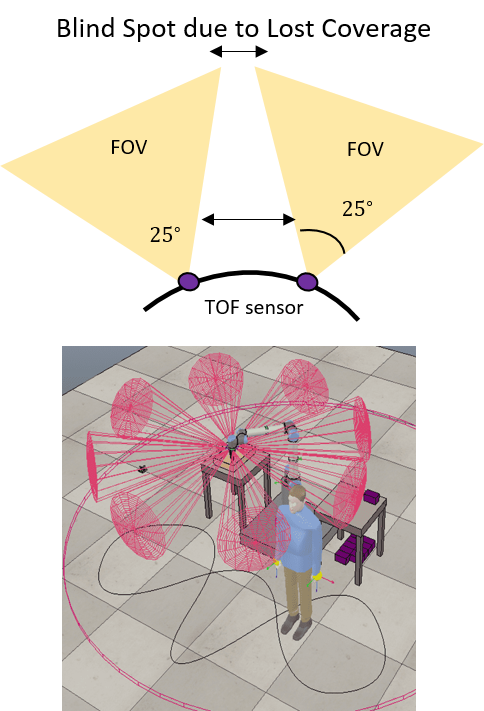}
			\label{fig:n1_8_0}}
	\end{subfigure} 
	\begin{subfigure}[]{
			\centering
		\includegraphics[width=0.4\linewidth,keepaspectratio]{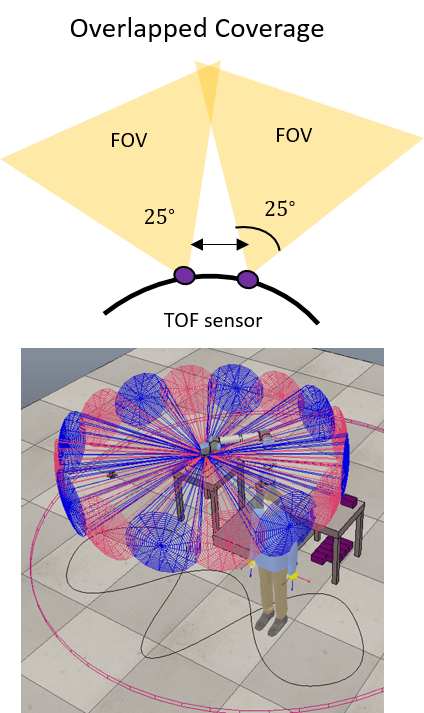}
			\label{fig:n1_16_0}}
	\end{subfigure}
	\caption{(a)The ToF Sensor Rings with 8 Sensor nodes i.e. \textit{n1\_8\_0}. There is a loss of coverage both far and near the robot. The simulated representation of the ToF ring mounted on the Tool is shown (bottom). (b) The ToF Sensor Rings with 16 Sensor nodes i.e. \textit{n1\_16\_0} showing overlapped coverage to compensate the lost coverage. }
	\label{fig:n1}
\end{figure}

In this work, we analyse the sparsity of ToF sensors per ring and the number and placement of rings on the robot links and its effect on the sensing volume coverage. Each ToF sensor in the ToF ring is a single unit solid-state lidar with a maximum detection range of $1.5m$ and a field-of-view (FOV) of $25deg$. The sensing volume of each sensor in a ToF ring can be represented as a cone of height $1.5m$ and angle $25^\circ$ degrees as shown in Fig \ref{fig:n1}. More details about the sensor setup and its use for safer HRC can be found in our previous works \cite{kumarDynamicAwarenessIndustrial2018} and \cite{CASE2019_paper}.

\begin{figure}[h!]
    \centering
    \includegraphics[width=8cm,keepaspectratio]{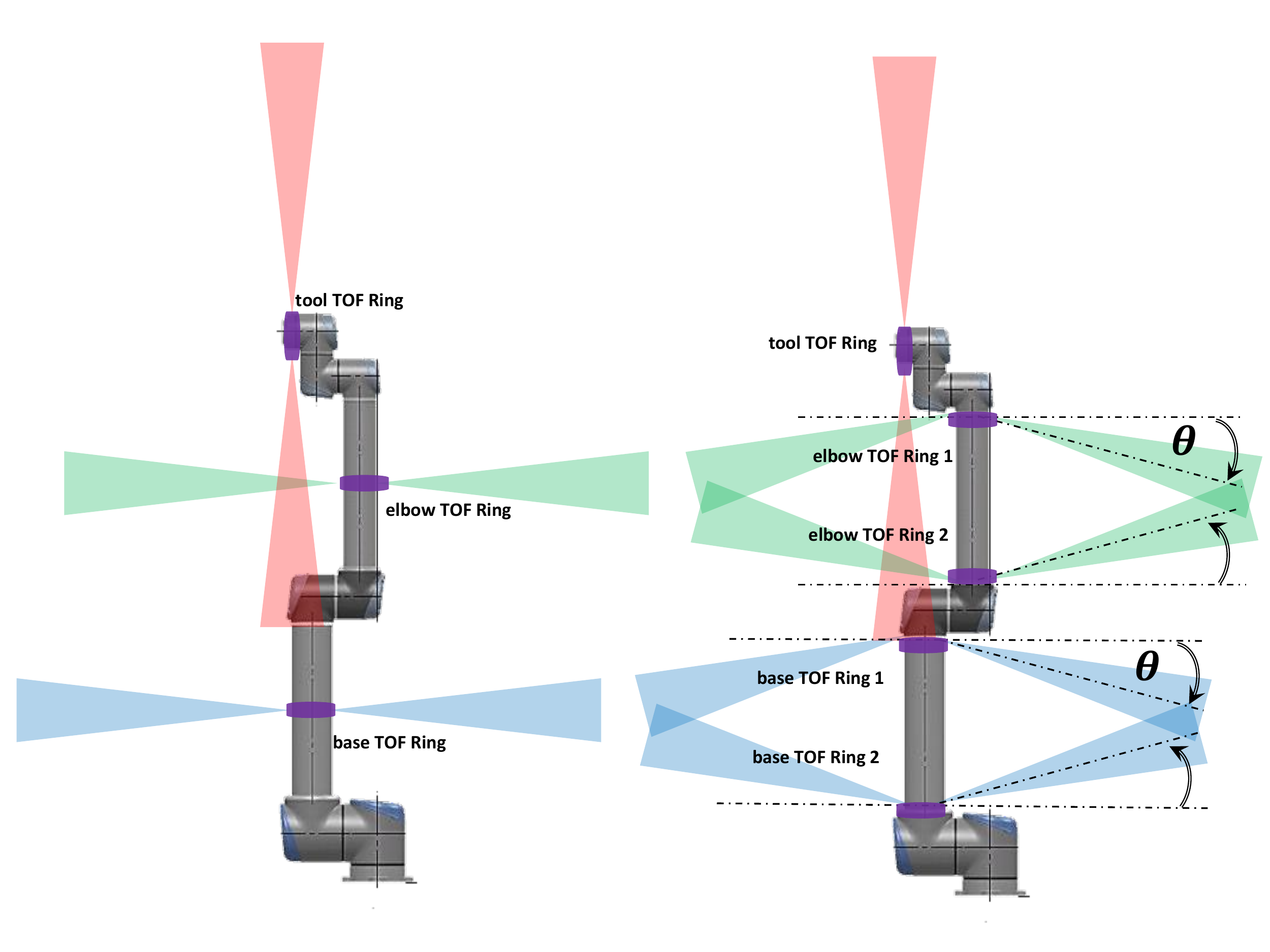}
    \caption{The sensor configuration for single  i.e. \textit{n1\_16\_0} and double rings i.e. \textit{n2\_16\_$\theta$}  (on shoulder and elbow links of UR 10) to measure sensing volume coverage. This configuration aims to cover volume near the robot.}
    \label{fig:n2}
\end{figure}
Different ToF sensor configurations are used to quantify the effect of blind-spots on sensing volume coverage. For brevity and ease of reference, different ToF sensor setup configurations as shown in Figure \ref{fig:n1} and \ref{fig:n2} are represented as follows: 
\begin{equation*}
\label{eq:sensor_config_rep}
 n \lbrace i \rbrace \text{\_}
\lbrace j \rbrace \text{\_}
\lbrace \theta \rbrace
\end{equation*}
\begin{equation*}
where 
\begin{cases}
    i& \text{ is num. of rings on the shoulder\&elbow links}\\
    j& \text{ is  num. of sensors per ring.}\\
    \theta & \text{ is the tilt angle of a sensor}\\
     & \text{w.r.t to the center of the ring.}
\end{cases}
\end{equation*}

\subsection{Sensing Volumes}
\label{sec:SensingVolumes}
\begin{figure}[h!]
    \centering
    \includegraphics[width=8.5cm,keepaspectratio]{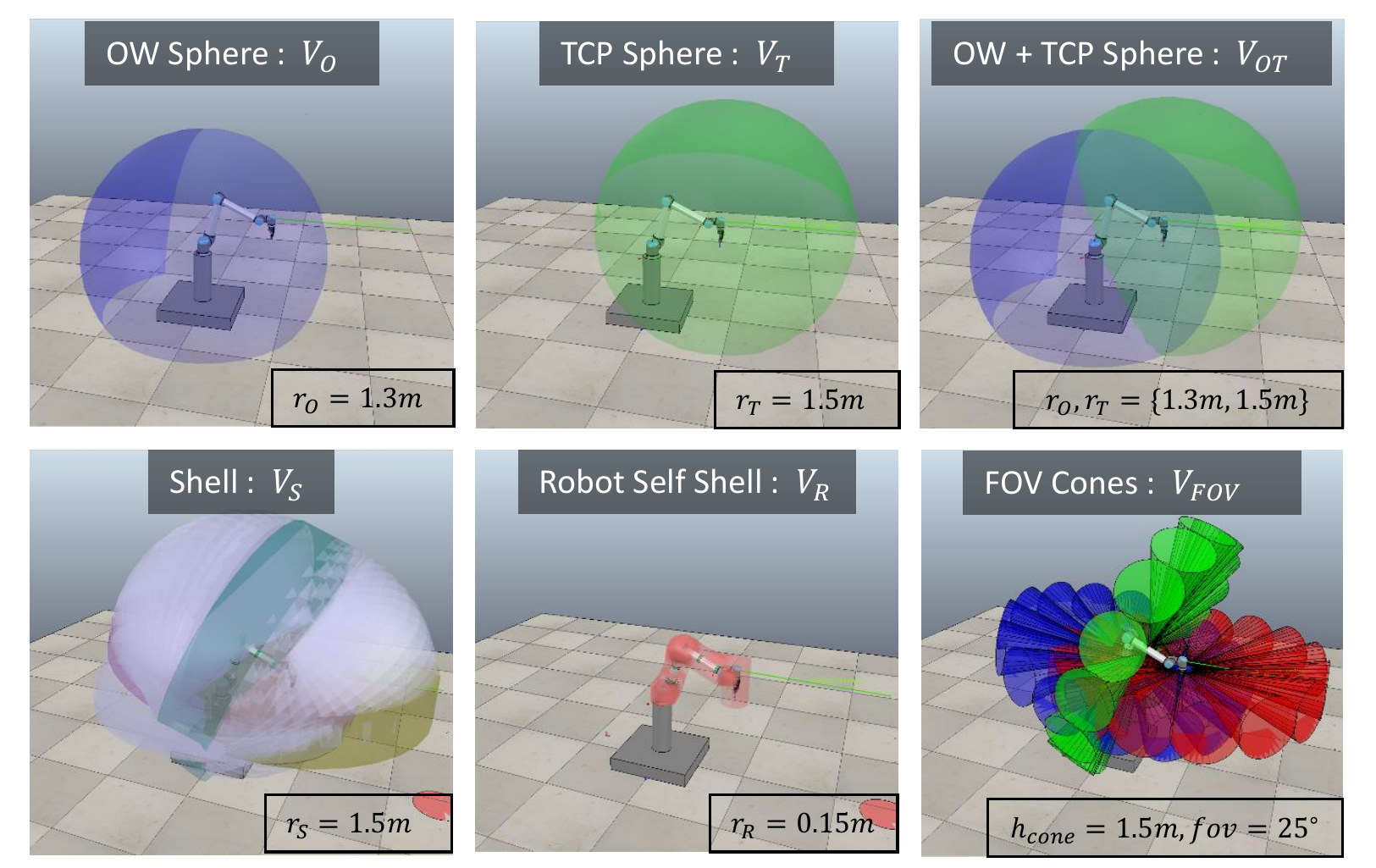}
    \caption{Maximum Sensing Volume $V_{max} = \lbrace V_O,V_T,V_{OT},V_S\rbrace$ used to quantify the sensing volume coverage of sensors with a total sensing volume coverage represented as $V_{FOV}$. The volume occupied by the robot self $V_R$, that is subtracted during Sensing Volume Coverage analysis.}
    \label{fig:MaxVolumes}
\end{figure}
Maximum Sensing Volume $V_{max}$ is the ideal workspace in Cartesian coordinate that the sensors should cover around the robot to ensure safe human robot interaction. Sensing volume coverage is the subset of this volume that is covered by the FOV of the ToF sensor rings. Maximum sensing volume differs based on the task, the application and the amount of human robot interaction.
In this work four different maximum volumes are suggested, and are shown in Figure \ref{fig:MaxVolumes}. They are described as follows: 
\begin{list}{\textbullet}{\leftmargin=0.5em}
    \item \textbf{Operating Workspace Volume} ($V_O$): This is the operating workspace of the robot. The robot used here is a UR 10 robot and its maximum reaching workspace is a sphere of radius $1.3m$. Sensing Volume Coverage of $V_O$ can be used to determine how much the ToF sensors cover near the robot. For tasks that require human close proximity to the robot, the ToF Setup that gives maximum coverage of this Volume can be considered.  
    \item \textbf{Tool (Tool Control Point -TCP) Volume } ($V_T$):  This is the sphere defined around the TCP of the robot $V_{T}$. Here the sphere radius is the maximum detection range of the ToF sensor i.e. $1.5m$. The TCP velocity and distance from TCP to human is mainly considered of safety in HRC \cite{marvelImplementingSpeedSeparation2017} \cite{ISOTS15066}. Hence, for scenarios where the robot performs Speed and Separation Monitoring (SSM) \cite{CASE2019_paper}, coverage in this volume space can be used to choose a ToF sensor configuration. 
    \item \textbf{Operating Workspace + Tool Volume} ($V_{OT}$): This is the combined volume in workspace. In order to determine the optimal ToF Sensor configuration that gives coverage for far and near volumes of the robot, the coverage in this combined volume can be used.
    \item \textbf{Shell Volume} ($V_S$): This is a tubular volume or a shell of fixed radius. The shell is defined along the curve comprised of the all robot link endpoints, with its starting point at the base link to the end at the TCP. This shell represents a more exact volume for which the sensing volume coverage should be maximized. 
\end{list}

\textbf{FOV Volume} for a sensor $j$ on ToF ring $j$ can be written as $^{ij} V_{fov}$. The combined sensing volume  $V_{FOV} = \sum (^{ij}V_{fov}) $ for  $\forall (i,j)$ (i.e. all ToF sensors in all the ToF rings). The overlap of this volume with the aforesaid volumes is used to determine the sensing coverage of a ToF sensor setup configuration.

\textbf{Inner Volume of the Robot} $V_R$ can be defined by the space occupied by the robot itself. Here we approximate it as a shell around the robot. In this work for UR 10 a shell of inner radius $0.15m$ is assumed (based on the maximum width of the bounding box of the largest link UR 10). This volume space is subtracted from all volumes.

\begin{figure}[h!]
    \centering
    \includegraphics[width=8cm,keepaspectratio]{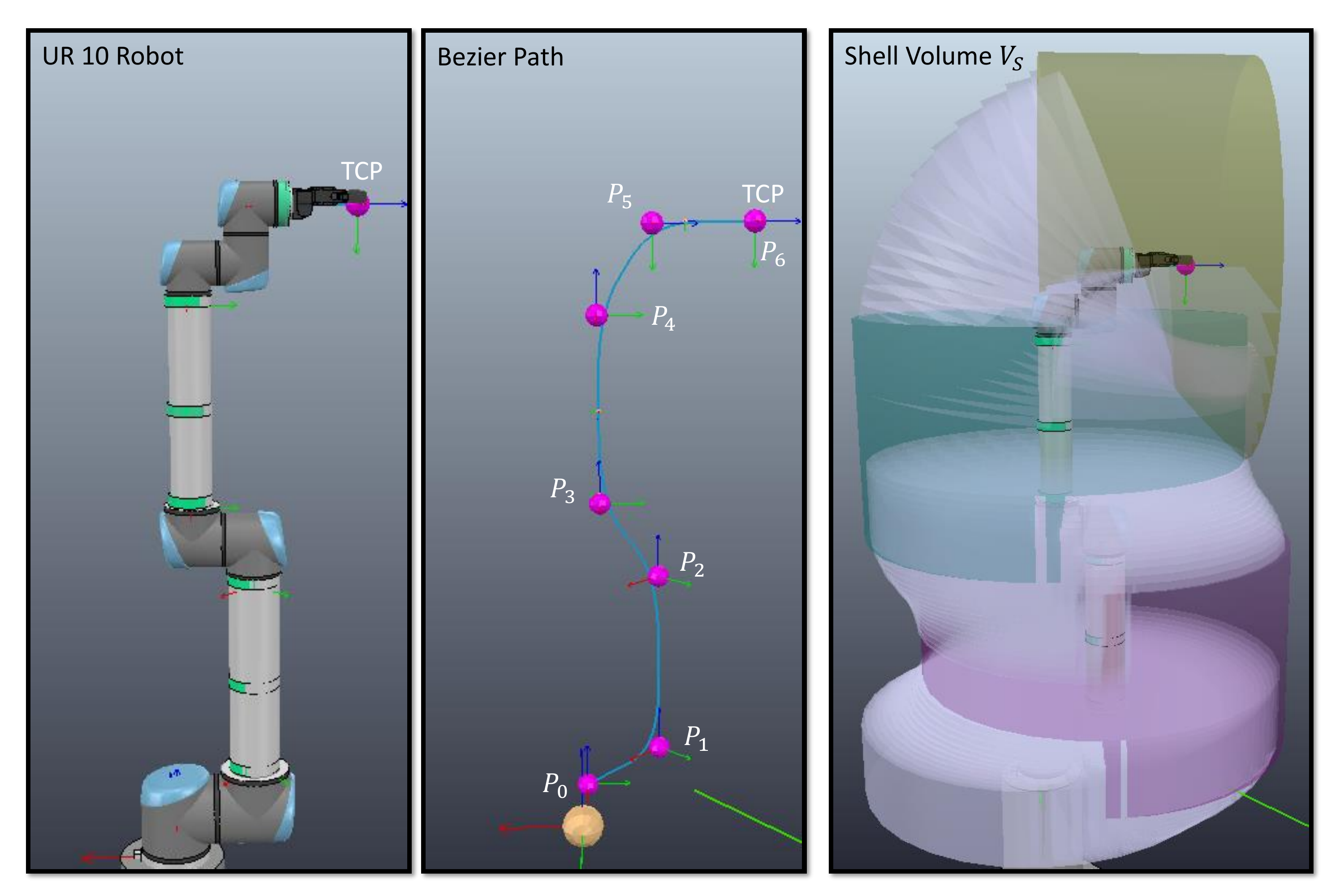}
    \caption{Generating Shell Volume (in image of radius $0.5m$) along the Bezier curve $r(t)$ generated by the UR10 robot link end points $P_0,P_1 .. P_6$. The Bezier interpolation is represented as the grey and where no interpolation was done is represented with differnt color.}
    \label{fig:Bezier_UR10}
\end{figure}
 In order to calculate the Shell Volume, Bezier interpolation of the robot pose using the end points of the robot links is done, refer Figure \ref{fig:Bezier_UR10}. This is detailed further in the following sections.

\subsubsection{Robot Pose as a Bezier curve}
\label{sec:BezierCurve}
For this work, a piece-wise Bezier curve approach has been used to generate a curve representing the robot pose. It essentially means part of the line segments between two points is not interpolated if it is co-linear, see  Figure \ref{fig:Bezier_UR10}. This was helpful as the interpolation was needed around the joints of the robot.  
\begin{figure}[h!]
    \centering
    \includegraphics[width=8cm,keepaspectratio]{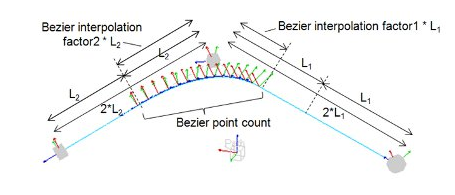}
    \caption{The Bezier Curve Interpolation for defining the curve given three control points $P_0,P_1 , P_2$ \cite{rohmerVREPVersatileScalable2013}. }
    \label{fig:bezier_curve_interpolation}
\end{figure}
A piece-wise Bezier interpolation determines at what point on the line segment between two control points does Bezier interpolation needs to be done. This is defined based on Bezier interpolation factors and the number of interpolation points (refer Figure \ref{fig:bezier_curve_interpolation}). The readers can refer to V-REP  API \cite{rohmerVREPVersatileScalable2013} and  \cite{BezierCurve1996} for more details.

\subsubsection{Shell Volume}
\label{sec:ShellVolume}
\begin{figure}[h!]
    \centering
    \includegraphics[width=8cm,keepaspectratio]{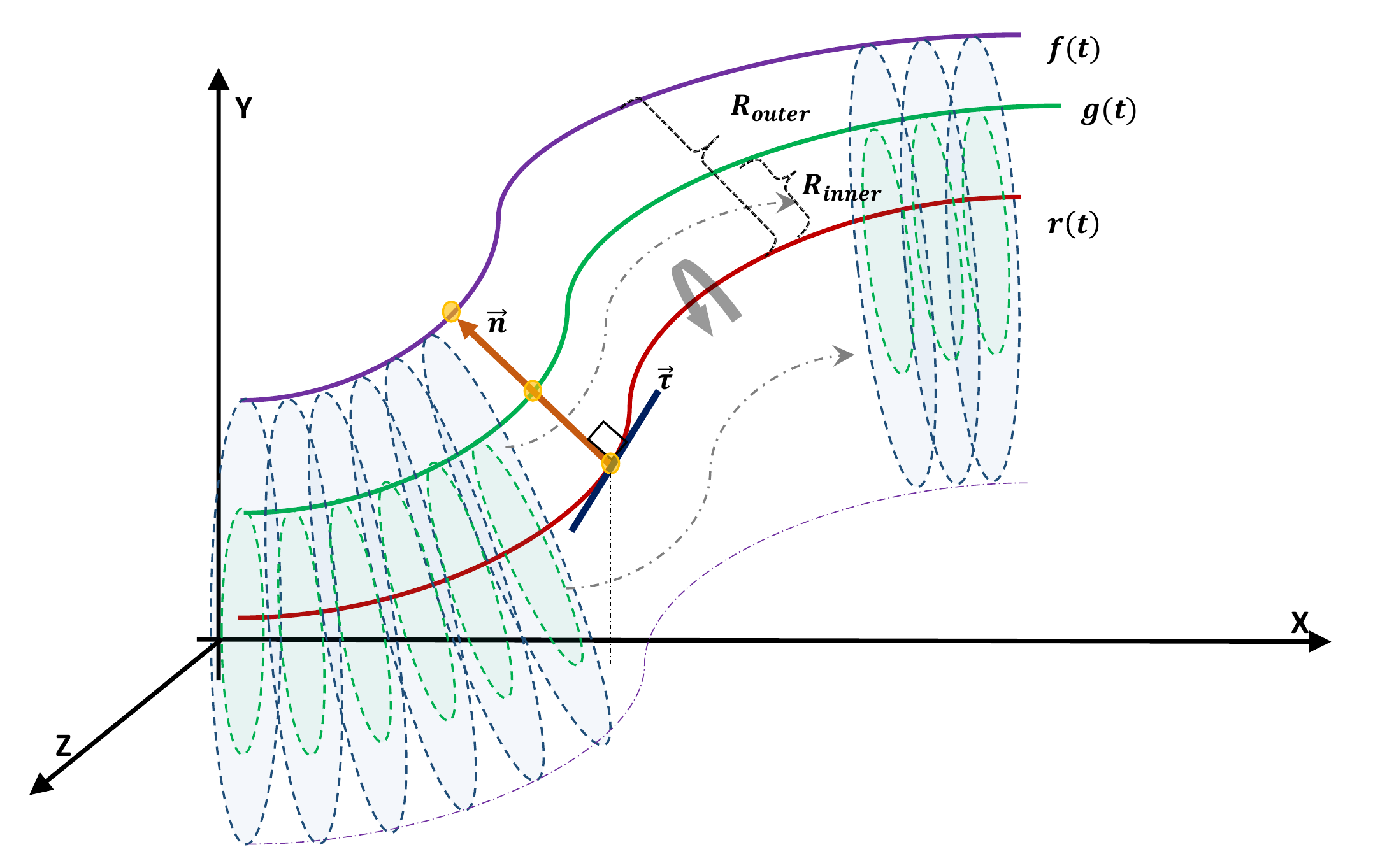}
    \caption{The shell volume calculation using the washer-method for curves $f(t)$ and $g(t)$ around $r(t)$ - Curved Axis Solids of Revolution.}
    \label{fig:graph_volume}
\end{figure}
This volume can be calculated by rotating a solid revolution along a curve. This is also know as `Curved Axis Solids of Revolution' \cite{CurvedAxis2013} (refer Figure \ref{fig:graph_volume}). This can be formulated as :
\begin{equation}
\begin{split}
    V_S = \pi \int_{t}^{} \left((\vec{f(t)}- \vec{r(t)})^2 - (\vec{g(t)}-\vec{r(t)})^2 \right)dt \\
    = \pi \int_{t}^{} \left( R_{outer} \bullet \vec{n(t)})^2 - (R_{inner} \bullet \vec{n(t)})^2 \right) dt \\
    \end{split}
\end{equation}
Alternatively according to Pappus Centroid Theorem \cite{PappusTheorem} as 
\begin{equation}
    V_S = \pi (R_{outer}^2 - R_{inner}^2) \bullet arclength(\vec{r(t)})
\end{equation}

where
\begin{subequations}
\begin{equation}
    \text{Length of Curve : } arclength(\vec{r(t)}) = \int_{t}^{} \lVert \vec{r(t)}'\rVert dt
\end{equation}
\begin{equation}
    \text{Outer Shell Vector :} \vec{f(t)} = \vec{r(t)} + R_{outer} \bullet \vec{n(t)} 
\end{equation}
\begin{equation}
    \text{Inner Shell Vector : }\vec{g(t)} = \vec{r(t)} + R_{inner} \bullet \vec{n(t)} 
\end{equation}
\end{subequations} 

\begin{subequations}
\begin{equation}
    \text{Normal Unit Vector : } \vec{n(t)} = \frac{\vec{\tau(t)}'}{\lVert \vec{\tau(t)}'\rVert}
\end{equation}
\begin{equation}
    \text{Tangent Unit Vector : }\vec{\tau(t)} = \frac{\vec{r(t)}'}{\lVert \vec{r(t)}'\rVert}
\end{equation}
\end{subequations}
Determining the volume covered by shell above can be computationally expensive, difficult to quantify especially with intersections of other volumes. Hence an approximation using Octree based volumetry has been done \cite{Octreeszeliski1993rapid}.
\subsubsection{Octree based Volumetry}
\label{sec:OctreeVolume}
\begin{figure}[htb]
    \centering
    \includegraphics[width=6.5cm,keepaspectratio]{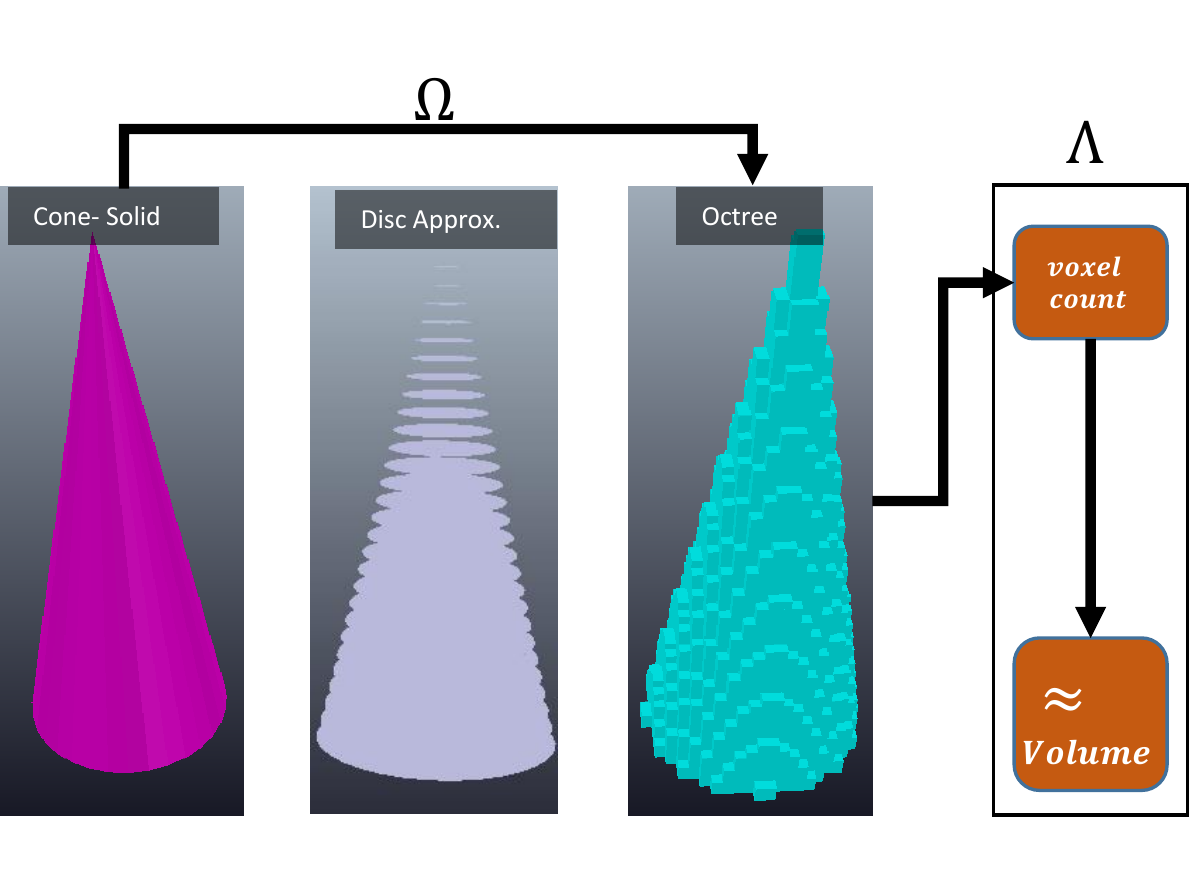}
    \caption{Octree based volumetry pipeline for a Cone shape. The $\Omega$ function represents a given shape as an octree and the $\Lambda$ operator quantifies the volume occupied by the octree. Volume of Cone is $0.173m^3$ and volume reported by Octree is $0.17m^3$. }
    \label{fig:octree_pipeline}
\end{figure}
Octrees are used for the voxel representation of any shape in a 3D Cartesian space. Octree representation of volume region $V_{max}$ is represented as $\Omega(V_{max})$. The volume can be calculated from octree by counting the number of voxels  in $\Omega(V_{max})$. A voxel is a cube with side length $l_{voxel}$ then the volume of the region occupied by octree $\Omega (V_{max})$ can be written as $\Lambda \left(\Omega(V_{max})\right) = voxelcount  \times l_{voxel}^3 $.  Octrees can be used to merge or subtract voxels from other octrees. Before a shape is converted into octrees, it is decimated into discs of varying radius spaced by voxel size $l_{voxel}$. This is done as the V-REP represents shapes as hollow and the inside volume of the shape needs to be calculated. An octree based volumetry pipeline for a FOV cone of a ToF sensor is shown in Figure \ref{fig:octree_pipeline}.

\subsection{Coverage of a ToF sensor configuration}
\label{sec:coveragevolume}
Given a maximum coverage volume $V_{max} = \lbrace V_{O},V_{T},V_{OT}, V_{S}\rbrace$, the coverage  $\zeta$, of a ToF sensor configuration ($n\_{i}\_{j}\_\theta^\circ$) with the field-of-view volume of $V_{FOV}$ can be written as: 
\begin{equation}
    \zeta (\%)= \frac{\Lambda \left(\Omega(V_{max}) \cap \Omega(V_{FOV})\right)}{\Lambda(\Omega(V_{max}))} \times 100
    \label{eq:Coverage of TOF Sensor1}  
\end{equation}
Alternatively as V-REP allows only addition and subtraction of voxels from Octrees, Equation \ref{eq:Coverage of TOF Sensor1} can be re-written as: 
\begin{equation}
   \zeta (\%)= \frac{\Lambda \left(\Omega(V_{max})\right) -
   \Lambda\left(\Omega(V_{max}) - \Omega(V_{FOV})\right)}{\Lambda(\Omega(V_{max}))} \times 100
   \label{eq:Coverage of TOF Sensor2} 
\end{equation}



\section{Experiments and Validation}
\label{sec:experiment_validation}
The experiment setup is a generic robot pick and place task of placing 10 products in a box (refer previous work \cite{kumarDynamicAwarenessIndustrial2018}, \cite{CASE2019_paper}). The robot movement involves moving the base joint $180^\circ$ degrees between the pick and place positions on the tables (refer Figure \ref{fig:n1}). This task was chosen as the base joint of a robot has the largest braking distance when moving at high joint speeds. This results in a radial motion of the Tool-Control-Point(TCP) i.e. the end-effector. 

In this work, the coverage at the robot pose when the robot is least safe i.e. moving at the highest speed during the task is measured. This was done with the reasoning that the ToF sensor arrays have the maximum coverage to detect and anticipate human/operator in the workspace. This setup is task specific but can be extended to any task which requires coverage either near or farther from the robot based on the human robot interaction during the task. Hence, in this study different $V_{max}$ volumes are considered that represent ideal maximum coverage both near and farther from the robot. For this study a UR 10 robot was simulated and the octree-based volume calculations were done using V-REP \cite{rohmerVREPVersatileScalable2013}.

In Speed and Separation Monitoring (SSM) based collaborative tasks \cite{marvelImplementingSpeedSeparation2017} the minimum distance calculations and directed velocities of human and robot are generally done with respect to the \textit{base} and \textit{TCP} of the robot. That is why the sensing volume coverage in a sphere representing the operating workspace ($V_{O}$) and a detection sphere centered at the TCP ($V_{T}$) are analyzed. However, according to the ISO standards \cite{ISOTS15066} and also in other works \cite{flaccoDepthSpaceApproach2015} the minimum distance and directed speeds can be w.r.t. any point on the robot. So a more exact sensing volume coverage is analyzed using a shell ($V_{S}$) around the robot self, that changes with the robot pose.

The sensing volume coverage is measured by determining the overlap of the ToF sensor arrays volume, $V_{FOV}$ with the maximum ideal volume $V_{max}$, which can be $V_{max} = \lbrace V_{O},V_T,V_{OT}, V_S \rbrace$ (refer Section \ref{sec:SensingVolumes}). 

\begin{figure}[h!]
    \centering
    \includegraphics[width=6.5cm,keepaspectratio]{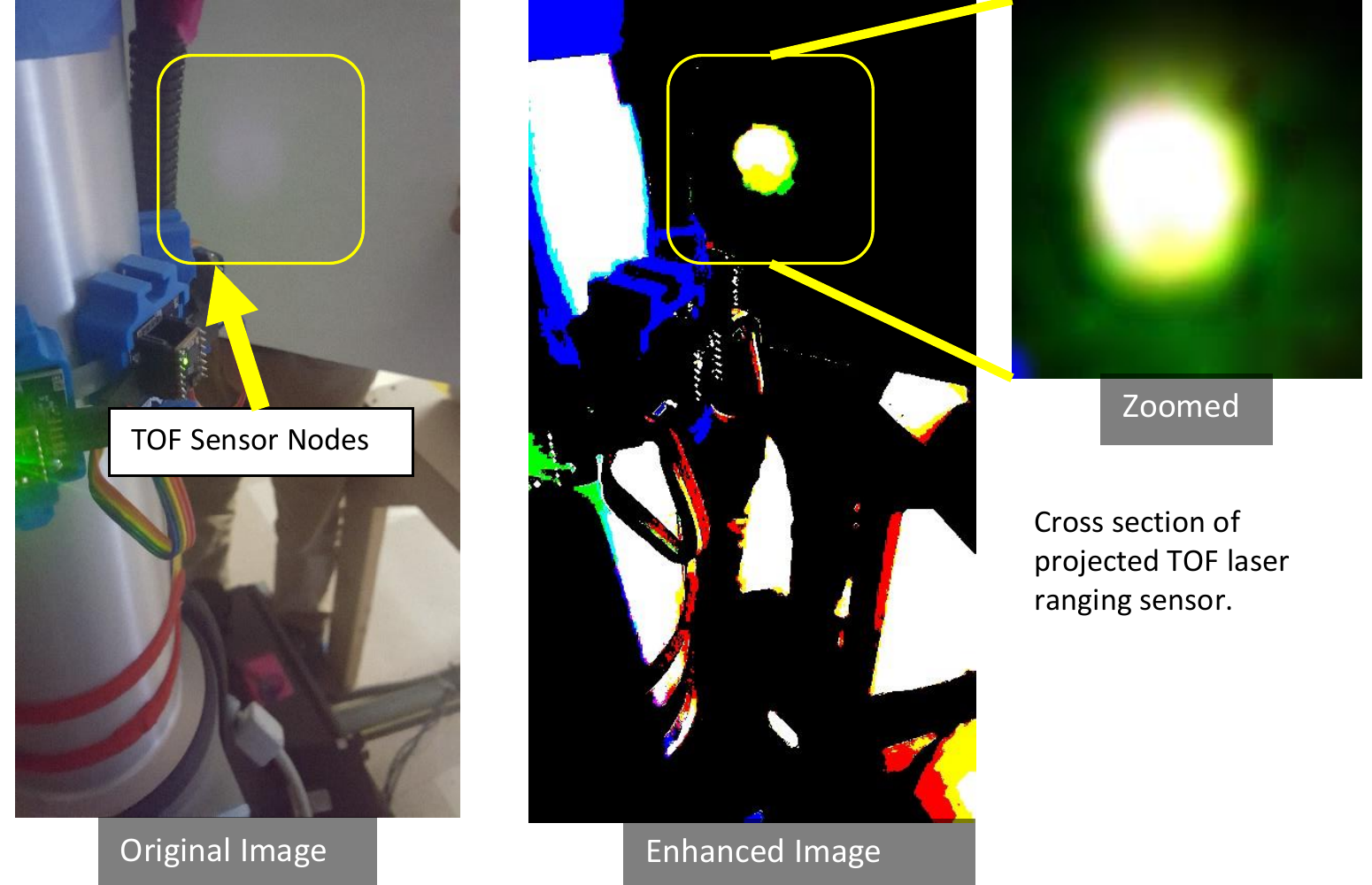}
    \caption{ Verification of the Time-of-Flight sensor node sensing volume can be modelled as a cone.}
    \label{fig:VolumeTOFProject}
\end{figure}
\subsection{Sensing Volume for ToF Sensor Arrays $V_{FOV}$}
\begin{figure}[h!]
    \centering
    \includegraphics[width=8cm,keepaspectratio]{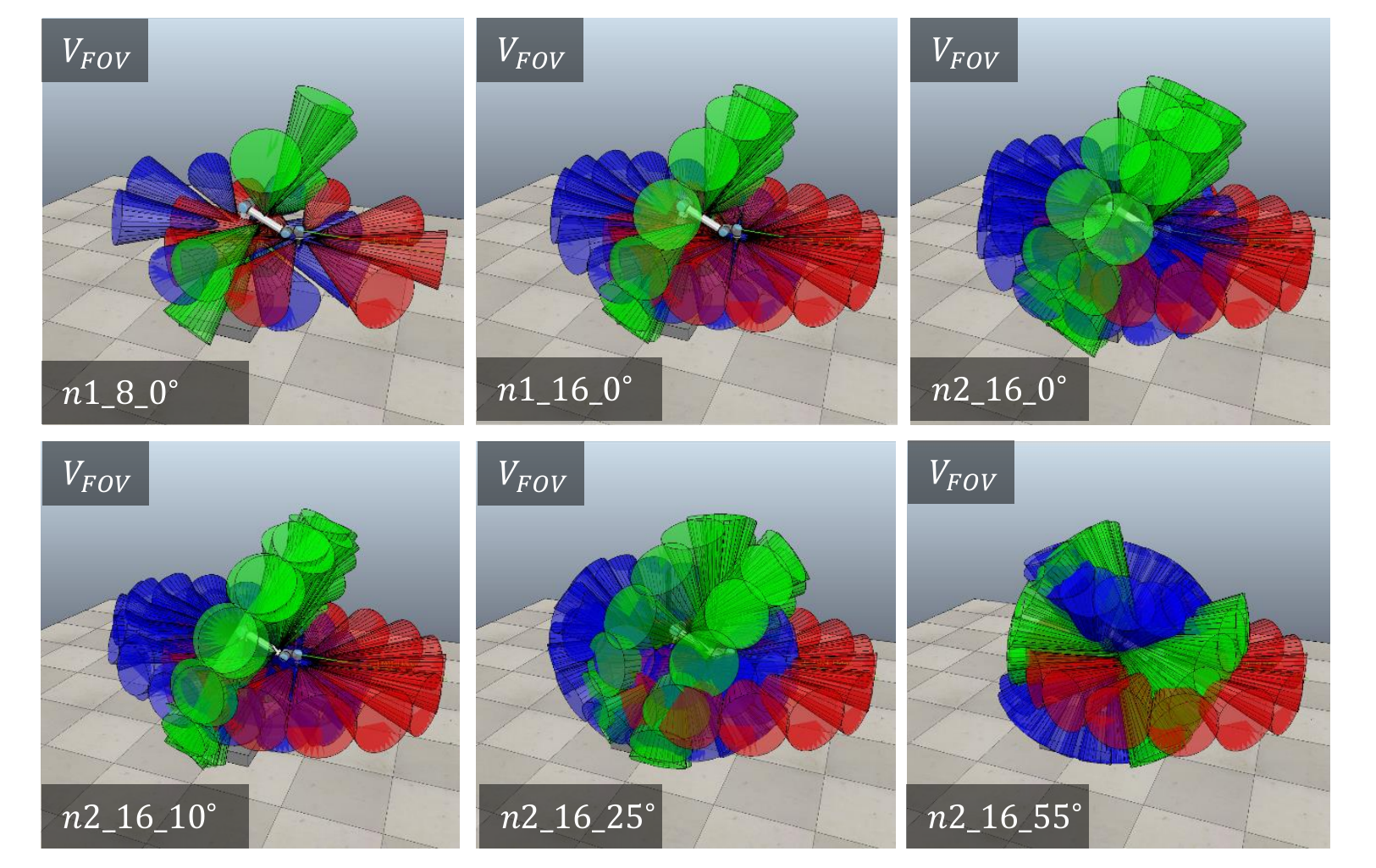}
    \caption{The ToF Sensor rings setups (Top Row) shows the 3 major ToF sensor configurations for $n1\_8\_0^\circ$ and $n1\_16\_0^\circ$  as single rings with 8 \& 16 sensors respectively and $n\_2\_16\_0^\circ$ - dual rings on shoulder and elbow links of the robot. The ToF Sensor rings setups (Bottom Row) shows the angle variation of sensors on the ring for $n2\_16\_\theta^\circ$ where $\theta^\circ \in \lbrace 0^\circ,25^\circ,55^\circ \rbrace$.  }
    \label{fig:VFOV_Setups}
\end{figure}

The sensor detection volume of a ToF laser ranging sensor is modelled as a cone with a field-of-view given by the beam angle of $25^\circ$ degrees and detection range i.e. the cone height as $1.5m$. In order to verify that the detection volume can be approximated as a cone a simple test of projecting the laser beam emitted by a ToF sensor on a white board was done and the image of the projection enhanced and the contour of the projection was checked. As shown in Figure \ref{fig:VolumeTOFProject} it can be seen that the projection shape approximates to a circle, which validates the modelling of ToF sensor detection volume as a cone. It can be seen in Figure \ref{fig:VFOV_Setups} the different $V_{FOV}$ sensing volumes for different ToF sensor configurations.
\subsection{Sensing Coverage Measurements}
\begin{figure}[h!]
    \centering
    \includegraphics[width=8cm,keepaspectratio]{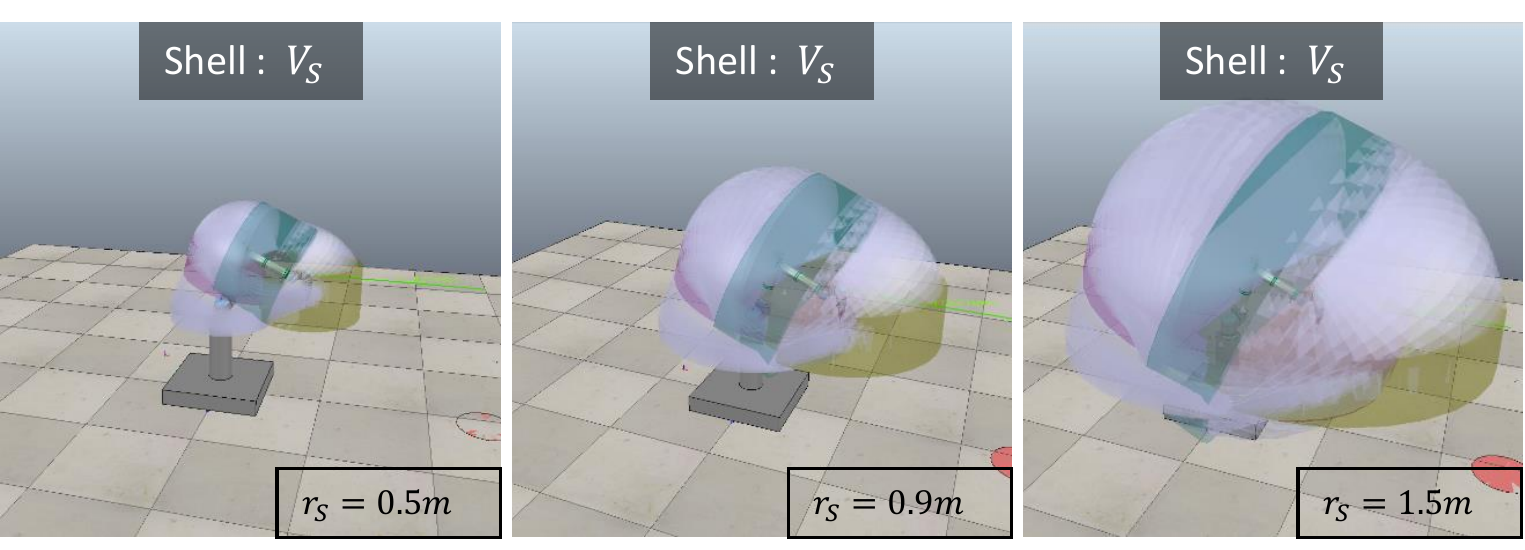}
    \caption{The ToF Sensor rings setups $n\_2\_16\_\theta^\circ$  are used to determine coverage for changing $V_S$ with varying radius $r_S$. The figure shown are $ r_S \in \lbrace 0.5m, 0.9m, 1.5m\rbrace$. }
    \label{fig:Shellradius}
\end{figure}

In order to analyse and compare the sensing volume coverage the following measurements were taken :
\begin{itemize}
    \item Impact on Sensing Coverage for ToF Configurations with different number of rings for all $V_{max}$ as shown in Figure \ref{fig:VFOV_Setups}-Top Row. The configurations compared were for single rings on elbow and shoulder robot links with 8 and 16 sensors per ring ($n1\_8\_0^\circ$ , $n1\_16\_0^\circ$), dual rings  with varying $\theta^{\circ} \in \lbrace10^{\circ},25^{\circ},55^{\circ} \rbrace$ ($n1\_16\_\theta^\circ$)and also three rings with ($n3\_16\_55^\circ$) which is mounting rings at the end of robot links at an angle $55^{\circ}$and also the center of the robot link.
    
    \item Sensing Coverage for ToF Configurations $n2\_16\_\theta^{\circ}$ with varying $\theta$ for all $V_{max}$. The $\theta^{\circ}$ is varied $5^{\circ}$ in the range of $\theta^{\circ} \in \left[0^\circ,60^\circ \right]$. This is to measure the impact of change in $\theta^{\circ}$ to the coverage in the near and farther zones of the robot.
    
    \item Sensing Coverage for ToF Configurations $n2\_16\_\theta^{\circ}$ with varying $\theta$ for shell volume $V_S$ with varying radius $r_S \in \lbrace1.5m,1.1m,0.9m,0.7m,0.5m\rbrace$ ( examples shown in Figure \ref{fig:Shellradius}).
\end{itemize}

 The results of these comparisons are shown and discussed in the following Section.


\section{Results}
\label{sec:Results}
\begin{figure}[]
    \centering
    \includegraphics[width=6.5cm,keepaspectratio]{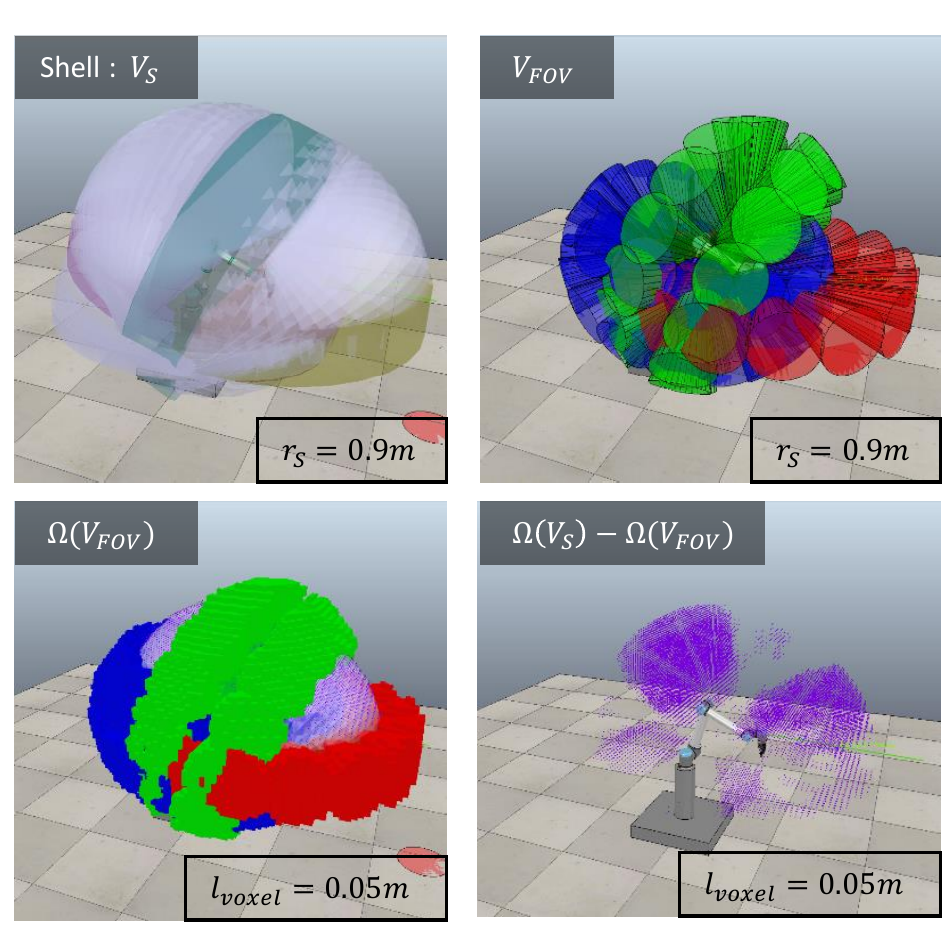}
    \caption{An example of Octree based approximation for calculating sensing volume coverage for shell volume $V_{max}=V_S$ of radius $r_S = 0.9m$ (Top Left). The $V_{FOV}$ of configuration $n2\_16\_25^\circ$ (Top Right) octree approximation(Bottom Left). The volume not covered by the sensors in the shell (Bottom Right).}
    \label{fig:Octree Example}
\end{figure}
Octree based approximations were used to calculate the sensing volumes $V_{max} \in \lbrace V_O,V_T,V_{OT},V_S \rbrace$ and also the ToF Sensor array volume $V_{FOV}$. A measurement for a shell $V_{max}=V_S$ of radius $r_S=0.9m$  is shown as an example in Figure \ref{fig:Octree Example}. In Figure \ref{fig:Octree Example}(Top Left) $V_S$ is shown, where the gray discs represent the Bezier interpolated points, where as the straight links are represented with other colors (refer Section \ref{sec:BezierCurve}). The $V_{FOV}$ for configuration $n2\_16\_25^{\circ}$ is shown in Figure \ref{fig:Octree Example}(Top Right), where red, green and blue cones represent the sensing volume of ToF rings mounted on tool, elbow and shoulder links of UR10 robot, respectively. The Octree approximation $\Omega(V_S)$ and $\Omega(V_{FOV})$ are shown in Figure\ref{fig:Octree Example}(Bottom Left). For visual clarity, $\Omega(V_S)$ is shown as a violet pointcloud where the points represent the center of the voxels in the octree. The left-over volume of $V_S$ not covered by the ToF rings is shown in Figure\ref{fig:Octree Example}(Bottom Right). Using Eq. \ref{eq:Coverage of TOF Sensor2} sensing volume coverage $\zeta(\%)$ was calculated (see Section \ref{sec:OctreeVolume}).

\begin{figure}[h!]
    \centering
    \includegraphics[width=8cm,keepaspectratio]{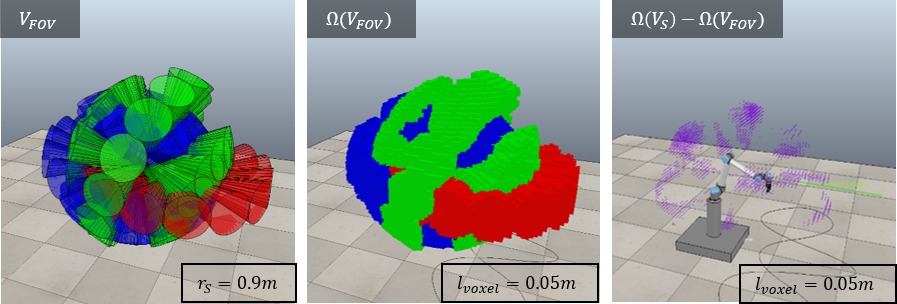}
    \caption{An Octree based approximation for calculating sensing volume coverage for shell volume $V_{max}=V_S$ of radius $r_S = 0.9m$  for configuration $n3\_16\_55^\circ$.}
    \label{fig:Octree_n3_16_55}
\end{figure}

\begin{figure}[h!]
    \centering
    \includegraphics[width=8cm,keepaspectratio]{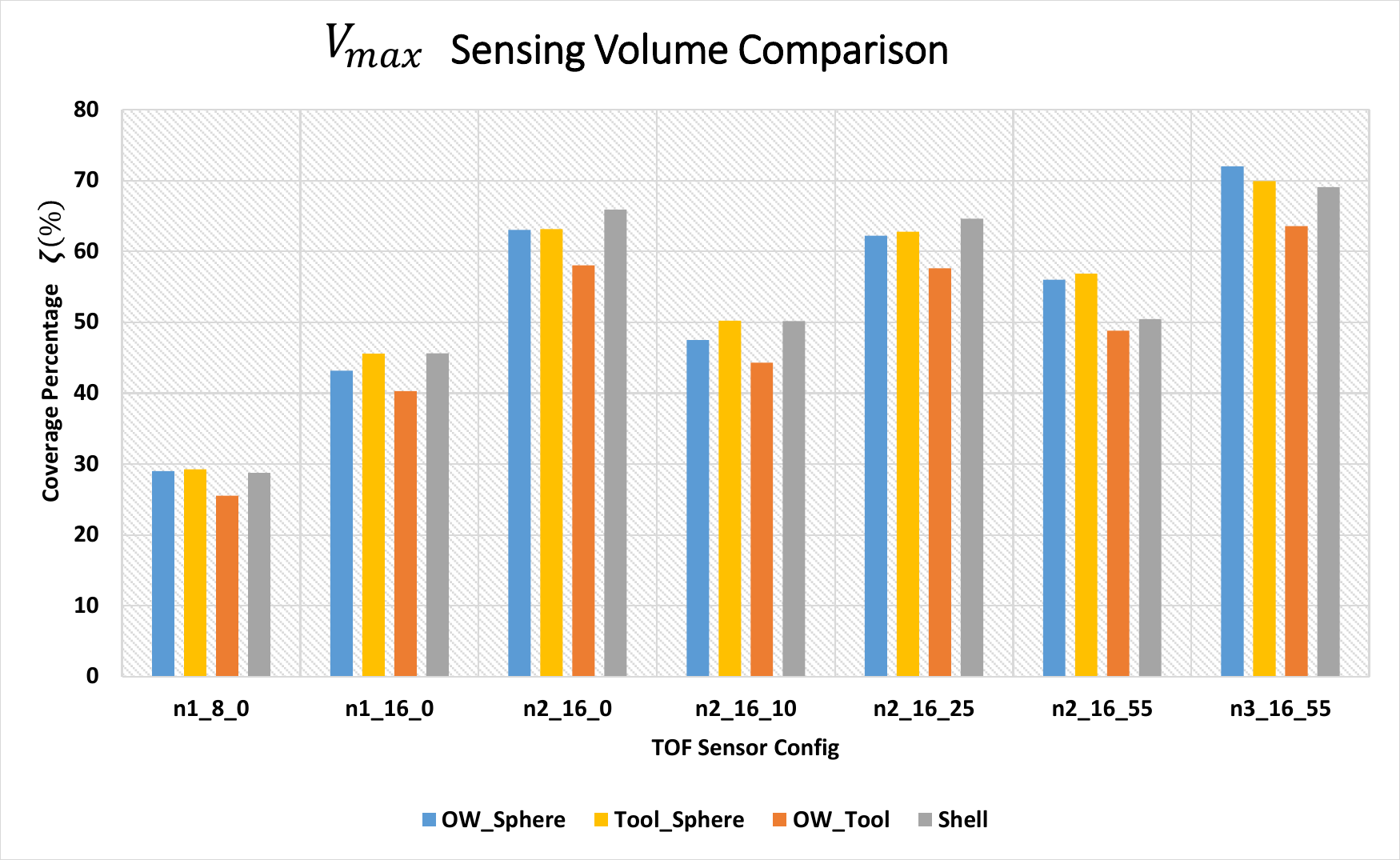}
    \caption{Sensing Volume Coverage $\zeta(\%)$ of ToF sensor configurations for all $V_{max}$ to observe impact of increasing number of sensors per ring i.e. $n1\_8\_0^{\circ}$ to $n1\_16\_0^{\circ}$ and increasing the number of rings per link i.e. $n1\_16\_0^{\circ}$, $n2\_16\_\theta^{\circ}$ and $n3\_16\_\theta^{\circ}$.}
    \label{fig:DiferentTOFConfigs}
\end{figure}

The first set of measurements were done for calculating $\zeta(\%)$ of ToF sensor configurations to observe impact of increasing number of sensors per ring i.e. $n1\_8\_0^{\circ}$ to $n1\_16\_0^{\circ}$ and increasing the number of rings per link i.e. $n1\_16\_0^{\circ}$, $n2\_16\_\theta^{\circ}$ and $n3\_16\_\theta^{\circ}$. The results are shown in the bar-graph in Figure \ref{fig:DiferentTOFConfigs}. The observations were as expected, an increasing coverage with more sensors per ring and more rings per link. Another observation that was made is that for $n2\_16\_10^{\circ}$ the coverage is similar to $n1\_16\_0^{\circ}$. This is because the the coverage of the two rings at $10^{\circ}$   overlap (as shown in Figure \ref{fig:VFOV_Setups}) and thus behave similar to a single ring in the center. Thus change in $\theta$ impacts the sensing volume coverage.

\begin{figure}[h!]
    \centering
    \includegraphics[width=8cm,keepaspectratio]{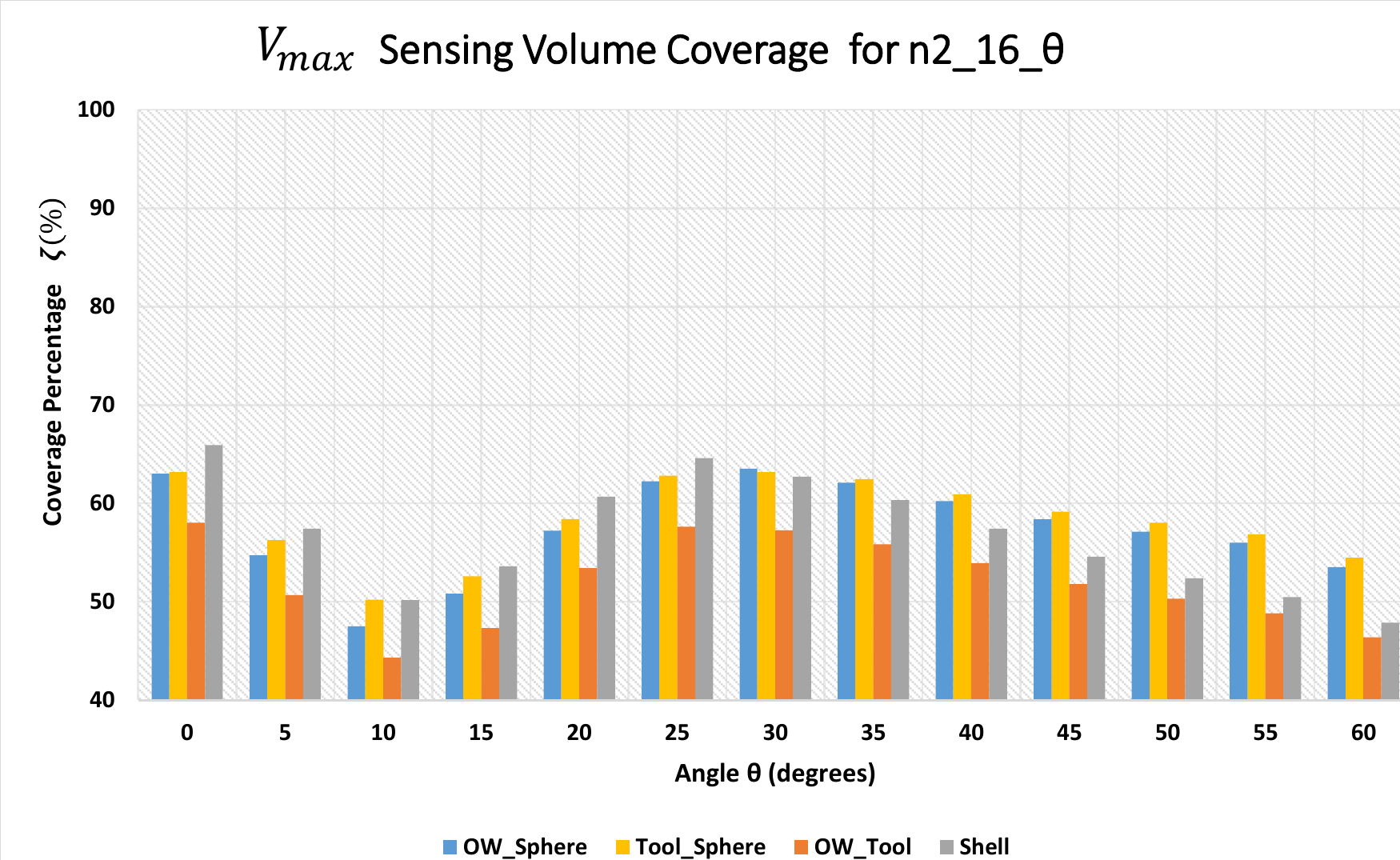}
    \caption{Sensing Volume Coverage $\zeta(\%)$ for all $V_{max}$ to observe impact of increasing $\theta$  in ToF sensor configurations  $n2\_16\_\theta^{\circ}$.}
    \label{fig:Result_All_n2_16_theta}
\end{figure}

To further observe the impact of change in $\theta$ in sensing volume coverage $\zeta(\%)$, $\theta$ is varied from $0^{\circ}$ to $60^{\circ}$ for the $n2\_16\_\theta^{\circ}$ ToF configuration. The results are shown in Figure \ref{fig:Result_All_n2_16_theta}. It is observed that as the overlap of the volume $V_{FOV}$ for a given set of ToF rings on a link increases, $\zeta(\%)$ drops. As observed before the coverage of $n2\_16\_10^{\circ}$ is minimum and equivalent to $n1\_16\_0^{\circ}$. It is observed that the most optimized and maximum coverage is given at $n2\_16\_0^{\circ}$ and $n2\_16\_25^{\circ}$ ToF sensor configuration.
\begin{figure}[h!]
    \centering
    \includegraphics[width=8cm,keepaspectratio]{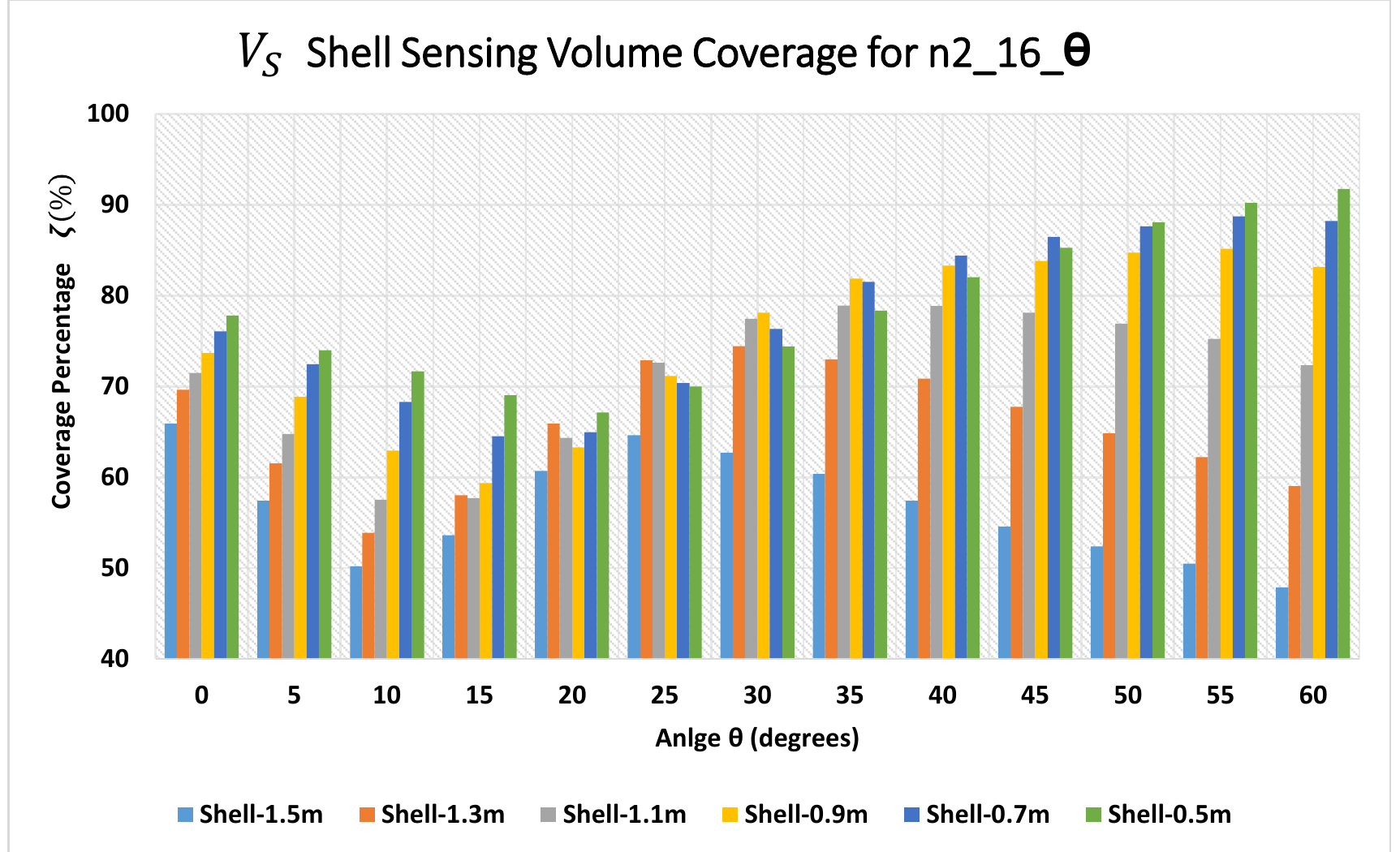}
    \caption{Sensing Volume Coverage $\zeta(\%)$ for shell volume $V_S$ with varying radius $r_S$  and varying $\theta$ in ToF sensor configurations  $n2\_16\_\theta^{\circ}$.}
    \label{fig:Result_Shell_n2_16_theta}
\end{figure}

In order to observe coverage in the range of $0.5m$ to $1.5m$ from the robot for varying $\theta$ in $n2\_16\_\theta^{\circ}$ ToF configuration, shell based volume $V_S$ of radius $r_S \in \lbrace0.5m,0.7m,0.9m,1.1m, 1.5m\rbrace$ was considered. In the previous work \cite{CASE2019_paper}, in the SSM implementation for safety using ToF sensors, $0.5m$ and $1.1m$ were considered as distance thresholds for varying the speeds of the robot. Thereby, if the human is working in close proximity, a sensor configuration that has more coverage closer to the robot can be used. Contrarily for farther distances and better anticipation of human encroaching on the robot workspace, farther coverage becomes important. Hence, sensing volume coverage $\zeta(\%)$ is calculated with varying $\theta$ and $r_S$. The results are shown in Figure \ref{fig:Result_Shell_n2_16_theta}. It can be observed that as $\theta$ increases the coverage near the robot also increases. It can be seen that for $r_S=0.5$ and $n2\_16\_55^{\circ}$, the coverage $\zeta > 90\% $.

In order to maximize the closer and father coverage, a sensor configuration that combines $n1\_16\_0^{\circ}$ and $n2\_16\_55^{\circ}$  i.e.  $n3\_16\_55^{\circ}$, which is placing three rings on the elbow and shoulder links of the robot is also implemented and the sensing volume coverage $\zeta(\%)$ measured. It results in a over $65\%$ coverage for all $V_{max}$ (shown in Figure \ref{fig:DiferentTOFConfigs}), and a coverage of $96.73\%$ and $69.80\%$ for $V_S$ with shell radius $r_S$ of $0.5m$ and $1.5m$ respectively. This is shown in Figure \ref{fig:Octree_n3_16_55}. The leftover $V_S$ can be compared to Figure \ref{fig:Octree Example}(Bottom Right) to see the difference in coverage.

\begin{figure}[h!]
    \centering
    \includegraphics[width=8cm,keepaspectratio]{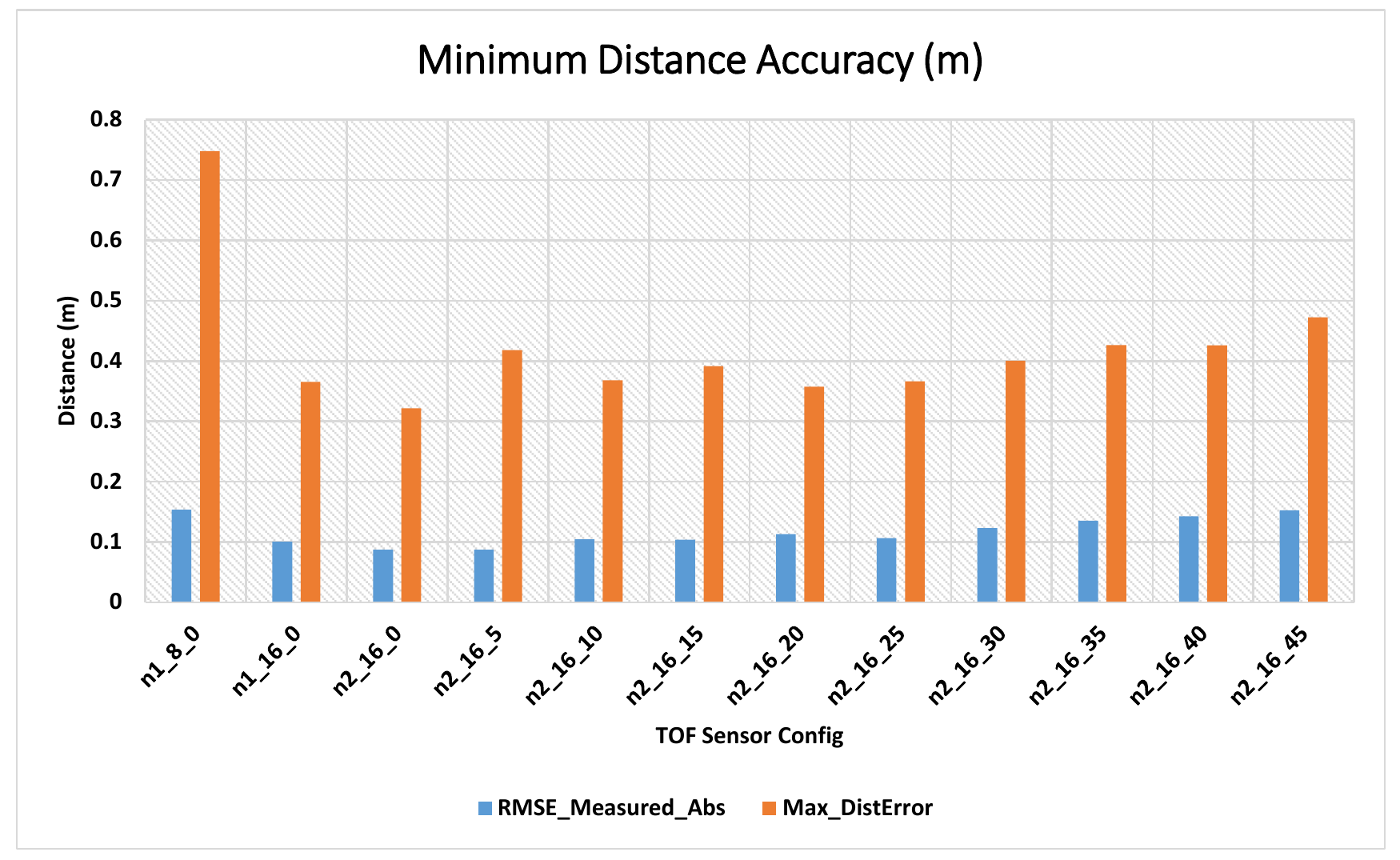}
    \caption{ Root Mean Square Error (RMSE) and the Maximum Distance Error between the measured minimum distance from the sensors between human-robot w.r.t. the absolute minimum distance i.e. the ground truth; for different ToF sensor configurations. }
    \label{fig:MinimumDistance}
\end{figure}

The minimum distance accuracy for a human moving in robot workspace for the experiment described in \cite{kumarDynamicAwarenessIndustrial2018} for the aforementioned ToF sensor configurations is shown in Figure \ref{fig:MinimumDistance}. Root Mean Square Error (RMSE) and the Maximum Distance Error between the measured minimum distance from the sensors between human-robot w.r.t. the absolute minimum distance i.e. the ground truth (the distance between the closest points on robot and the human) is shown. It was observed that as the sensing volume coverage increases with the number of sensors per ring and number of rings per link the error decreases.   




\section{Conclusion}
\label{sec:Conclusion}
 In this paper, a methodology to quantify sensing volume coverage of on-robot Time-of-Flight laser ranging sensor arrays/rings was presented. The measurements of this sensing volume coverage were presented for various ToF sensor configurations. It was observed that for the configuration of dual rings mounted on robot links with an angle of $25^{\circ}$ and $0^{\circ}$ degrees gave the optimal coverage for closer and farther regions. Increasing the number of sensor per ring and reducing the blind-spots increases the coverage and minimum distance accuracy and increasing the number of rings per link also helps with the sensing volume coverage of the robot.
 
 The current ongoing work is validating the affect on minimum distance accuracy for these configurations with different sized objects and human in the workspace in the closer and farther zones of the robot workspace.







\section*{Acknowledgment}
The authors are grateful to the staff of Multi Agent Bio-Robotics Laboratory (MABL) and the CM Collaborative Robotics Research (CMCR) Lab for their valuable inputs.
\bibliographystyle{IEEEtran}
\bibliography{papers_v1}{}

\end{document}